\documentclass[runningheads]{llncs}

\usepackage{eccv}

\usepackage{eccvabbrv}

\usepackage[dvipsnames]{xcolor}

\usepackage{graphicx}
\usepackage{amsmath}

\usepackage{amssymb}
\usepackage{wasysym}
\usepackage{booktabs}

\usepackage{bm}
\usepackage{algorithm}
\usepackage{algpseudocode} \usepackage{multirow}
\usepackage{wrapfig}
\usepackage{caption}
\usepackage{listings}
\usepackage{algpseudocode}\usepackage[T1]{fontenc}
\usepackage{xcolor}
\usepackage{colortbl}

\newlength\savewidth\newcommand\shline{\noalign{\global\savewidth\arrayrulewidth
\global\arrayrulewidth 1pt}\hline\noalign{\global\arrayrulewidth\savewidth}}

\usepackage{array}
\renewcommand{\paragraph}[1]{\noindent\textbf{#1}}

\newcommand{\tablestyle}[2]{\setlength{\tabcolsep}{#1}\renewcommand{\arraystretch}{#2}\centering\footnotesize}

\definecolor{defaultcolor}{gray}{.9}

\definecolor{ourscolor}{HTML}{EFFAF1}

\definecolor{deemph}{gray}{0.6}
\newcommand{\gc}[1]{\textcolor{deemph}{#1}}
\definecolor{decoded_color}{HTML}{A5DCE0}
\definecolor{masked_color}{HTML}{A7A9AC}

\definecolor{ratio}{rgb}{0.0, 0.0, 0.0}
\definecolor{sampt}{rgb}{0.0, 0.0, 0.0}
\definecolor{remaskt}{rgb}{0.0, 0.0, 0.0}
\definecolor{cfg}{rgb}{0.0, 0.0, 0.0}

\usepackage{pifont}

\usepackage{lipsum}

\newcommand{\pub}[1]{\xspace\textcolor{gray}{\tiny{(\textit{#1})}}}
\definecolor{Highlight}{HTML}{39b54a}  \newcommand{\hl}[1]{\textcolor{Highlight}{\textbf{#1}}}

\usepackage{xspace}
\newcommand{\VQVAE}{VQVAE-2$^\ddagger$~\cite{razavi2019generating}\pub{NeurIPS'19}\xspace}

\newcommand{\LDM}{LDM~\cite{rombach2022high}\pub{CVPR'22}\xspace}

\newcommand{\ADMp}{ADM-G$^\dagger$~\cite{dhariwal2021diffusion}\pub{NeurIPS'21}\xspace}
\newcommand{\LDMp}{LDM$^\dagger$~\cite{rombach2022high}\pub{CVPR'22}\xspace}
\newcommand{\UVITp}{U-ViT-H$^\dagger$~\cite{bao2022all}\pub{CVPR'23}\xspace}
\newcommand{\DITp}{DiT-XL$^\dagger$~\cite{peebles2023scalable}\pub{ICCV'23}\xspace}

\newcommand{\VQGAN}{VQGAN$^\ddagger$~\cite{esser2021taming}\pub{CVPR'21}\xspace}
\newcommand{\VQDiffusion}{VQ-Diffusion$^\ddagger$~\cite{gu2022vector}\pub{CVPR'22}\xspace}
\newcommand{\MaskGIT}{MaskGIT$^\ddagger$~\cite{chang2022maskgit}\pub{CVPR'22}\xspace}

\newcommand{\TokenCritic}{Token-Critic$^\ddagger$~\cite{lezama2022improved}\pub{ECCV'22}\xspace}
\newcommand{\DraftRevise}{Draft-and-revise~\cite{lee2022draft}\pub{NeurIPS'22}\xspace}
\newcommand{\MAGE}{MAGE$^\ddagger$~\cite{li2023mage}\pub{CVPR'23}\xspace}
\newcommand{\MaskgitCfg}{MaskGIT~\cite{chang2022maskgit}\pub{CVPR'22}\xspace}
\newcommand{\MaskgitFSQ}{MaskGIT-FSQ~\cite{mentzer2023finite}\pub{ICLR'24}\xspace}
\newcommand{\Ours}{AdaNAT\xspace}

\newcommand{\USF}{USF~\cite{liu2023unified}\pub{ICLR'24}\xspace}

\newcolumntype{x}[1]{>{\centering\arraybackslash}p{#1pt}}
\newcolumntype{y}[1]{>{\raggedright\arraybackslash}p{#1pt}}
\newcolumntype{z}[1]{>{\raggedleft\arraybackslash}p{#1pt}}

\newcommand{\learnable}{\textcolor{learnable}{\textsf{\textit{{learnable}}}}\xspace}
\newcommand{\adaptive}{\textcolor{adaptive}{\textsf{\textit{{adaptive}}}}\xspace}
\definecolor{existing}{HTML}{7F7F7F} \definecolor{learnable}{HTML}{C00000} \definecolor{adaptive}{HTML}{0070C0} \definecolor{imagereward}{HTML}{C00000} \definecolor{agent}{HTML}{1C5CB0} \definecolor{agentcolor}{HTML}{7AACE9} \definecolor{rmcolor}{HTML}{EB8253} \definecolor{frozen}{HTML}{6F8EB6}

\usepackage{empheq}

\newcommand{\cmark}{\ding{51}}\newcommand{\xmark}{\ding{55}}\usepackage{enumitem}

\usepackage{cases}
\usepackage{graphicx}
\usepackage{booktabs}

\usepackage[accsupp]{axessibility}

\usepackage{hyperref}

\usepackage{orcidlink}

\begin{document}

\title{AdaNAT: Exploring Adaptive Policy for Token-Based Image Generation}

\titlerunning{AdaNAT}

\author{Zanlin Ni\inst{1}\thanks{Equal contribution.\ \ \ \ \ $^\dagger$Corresponding Authors.}\orcidlink{0009-0006-6428-1456} \and
Yulin Wang\inst{1}\inst{\star}\orcidlink{0000-0002-1363-0234} \and
Renping Zhou\inst{1}\orcidlink{0009-0002-0546-0885} \and
Rui Lu\inst{1}\orcidlink{0009-0003-4850-8401} \and \\
Jiayi Guo\inst{1}\orcidlink{0009-0005-7004-939X} \and
Jinyi Hu\inst{1}\orcidlink{0009-0002-9440-4198} \and
Zhiyuan Liu\inst{1}\orcidlink{0000-0002-7709-2543} \and
Yuan Yao\inst{2}{$^\dagger$}\orcidlink{0000-0002-8276-3620} \and
Gao Huang\inst{1}{$^\dagger$}\orcidlink{0000-0002-7251-0988}
}

\authorrunning{Z. Ni \emph{et al.}}

\institute{
    \small {Tsinghua University \and National University of Singapore}
}

\maketitle

\begin{abstract}
Recent studies have demonstrated the effectiveness of token-based methods for visual content generation.
As a representative work, non-autoregressive Transformers (NATs) are able to synthesize images with decent quality in a small number of steps.
However, NATs usually necessitate configuring a complicated generation policy comprising multiple manually-designed scheduling rules.
These heuristic-driven rules are prone to sub-optimality and come with the requirements of expert knowledge and labor-intensive efforts.
Moreover, their one-size-fits-all nature cannot flexibly adapt to the diverse characteristics of each individual sample.
To address these issues, we propose \Ours, a \emph{learnable} approach that automatically configures a suitable policy \emph{tailored for} every sample to be generated.
In specific, we formulate the determination of generation policies as a Markov decision process.
Under this framework, a lightweight policy network for generation can be learned via reinforcement learning.
Importantly, we demonstrate that simple reward designs such as FID or pre-trained reward models, may not reliably guarantee the desired quality or diversity of generated samples.
Therefore, we propose an adversarial reward design to guide the training of policy networks effectively.
Comprehensive experiments on four benchmark datasets, \ie, ImageNet-256$^2$\&512$^2$, MS-COCO, and CC3M, validate the effectiveness of \Ours.
Code and pre-trained models will be released at \url{https://github.com/LeapLabTHU/AdaNAT}.
\keywords{non-autoregressive Transformers \and reinforcement learning \and adaptive image generation}
\end{abstract}
    
\section{Introduction}
\label{sec:introduction}

\begin{figure}[t]
    \centering
    \resizebox{.95\columnwidth}{!}{     \begin{minipage}{.36\columnwidth}
        \centering
        \includegraphics[width=\linewidth]{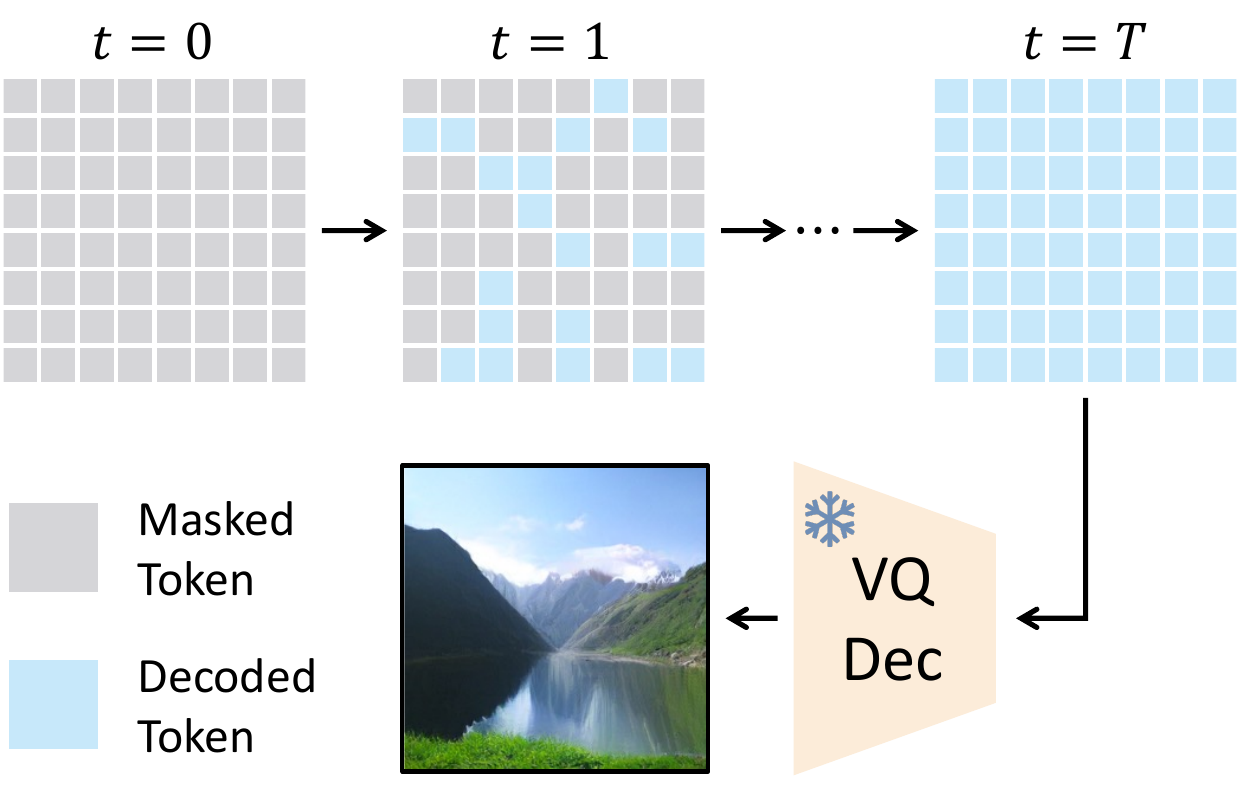}
        \caption*{
            (a)
        }
        \resizebox{\columnwidth}{!}{
            \tablestyle{4pt}{1.05}
            \begin{tabular}{x{57}|x{40}x{40}}
                & \multicolumn{2}{c}{generation policy} \\
                & \emph{learnable?} & \emph{adaptive?} \\\shline
                \multirow{2}{*}{\shortstack{existing works\\~\cite{chang2022maskgit,lezama2022improved,li2023mage,chang2023muse,yu2023language}}} & \multirow{2}{*}{\xmark} & \multirow{2}{*}{\xmark} \\
                &            \\
                \textbf{AdaNAT} & \cmark            & \cmark
            \end{tabular}
        }
        \caption*{
            (b)
            \label{tab:comp_existing_with_ours}
        }
    \end{minipage}
    \begin{minipage}{0.64\columnwidth}
            \centering
            \includegraphics[width=\linewidth]{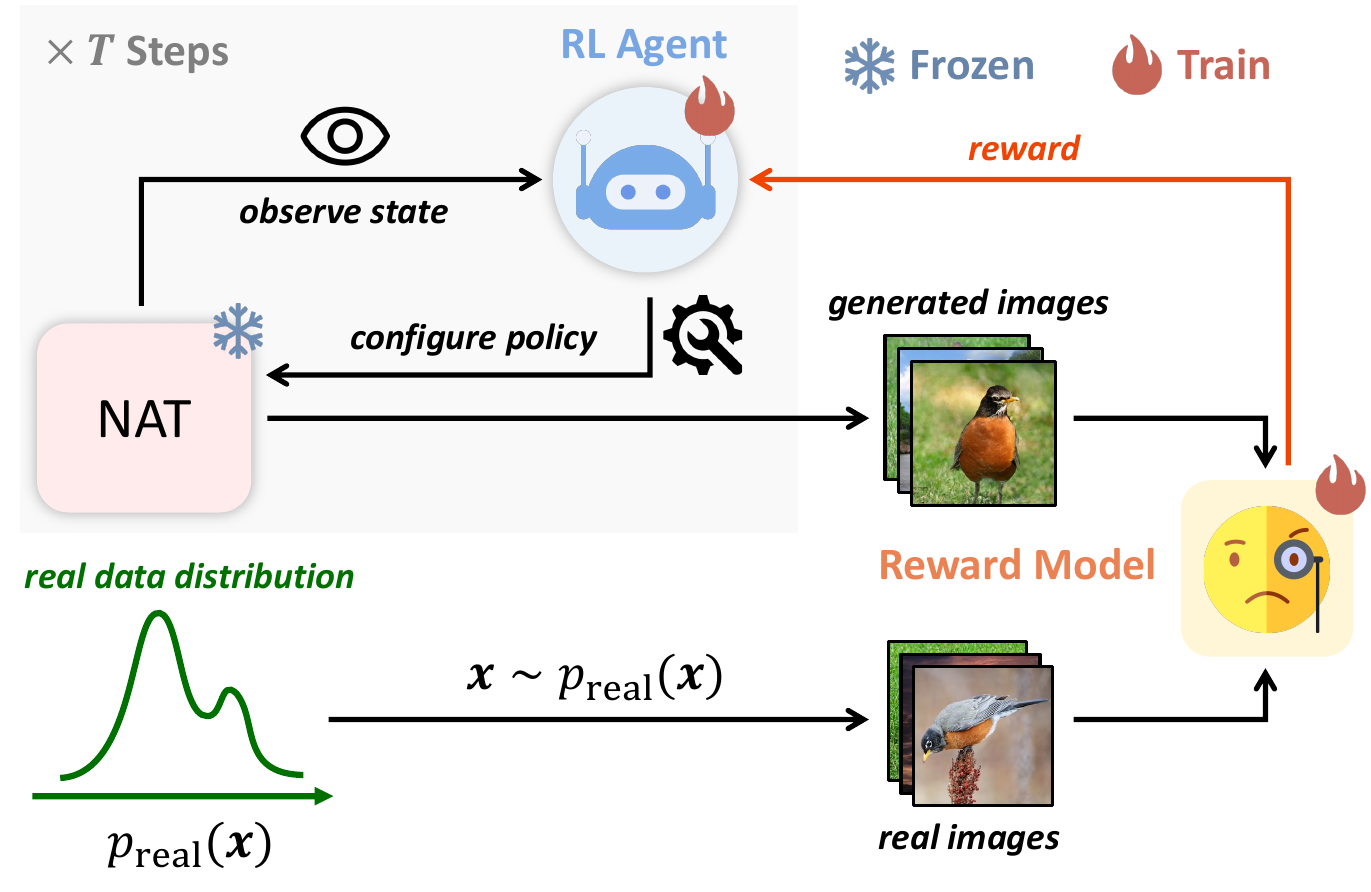}
            \caption*{
                (c)
            }
    \end{minipage}
    }
    \caption{
        (a) \textbf{The generation process of non-autoregressive Transformers (NATs)} starts from a entirely masked canvas and parallely decodes multiple tokens at each step.
        The fully decoded tokens are then mapped to the pixel space with a pre-trained VQ-decoder~\cite{esser2021taming}.
        (b) \textbf{Existing works \vs \Ours.}
        (c) \textbf{Overview of \Ours with adversarial reward modeling.}
        The \textcolor{agentcolor}{\textbf{RL agent}} is our policy network, which configures the suitable policy based on the observed generation status.
        The \textcolor{rmcolor}{\textbf{reward model}}, providing the probability of a sample being real as the reward, is simultaneously refined to better discriminate between real and fake samples.
        It is important to note that \Ours only learns the proper generation policy upon a pre-trained NAT model.
        The NAT model itself is kept \textcolor{frozen}{\textbf{frozen}} throughout our pipeline.
        \label{fig:teaser}
    }
\end{figure}

Recent years have witnessed unprecedented growth in the field of AI-generated content (AIGC).
In computer vision, diffusion models~\cite{dhariwal2021diffusion,rombach2022high,saharia2022photorealistic} have emerged as an effective approach.
On the contrary, within the context of natural language processing, content is typically synthesized via the generation of discrete tokens using Transformers~\cite{vaswani2017attention,ghazvininejad2019mask,brown2020language,raffel2020exploring}.
Inspired by such discrepancy, there has been a growing interest in investigating this token-based generation paradigm for visual synthesis~\cite{chang2022maskgit,yu2022scaling,lezama2022improved,yu2023scaling,chang2023muse,li2023mage}.
Different from diffusion models, these approaches utilize a discrete data format akin to language models.
This makes them straightforward to harness well-established language model optimizations such as refined scaling strategies~\cite{brown2020language,rae2021scaling,hoffmann2022training,wang2022language} and advances in model infrastructure~\cite{shoeybi2019megatron,du2022glam,chowdhery2023palm}.
Moreover, explorations in this field may facilitate the development of more advanced, scalable multimodal models with a shared token space~\cite{ge2023planting,team2023gemini,ge2023making,mizrahi20244m,zhan2024anygpt} as well as general-purpose vision foundation models that unify visual understanding and generation capabilities~\cite{li2023mage,tian2023addp}.

In the direction of token-based visual generation, the recently proposed non-autoregressive Transformers (NATs) have exhibited noteworthy potential in terms of both generation quality and computational efficiency~\cite{chang2022maskgit,lezama2022improved,chang2023muse,qian2023strait}.
Compared to traditional autoregressive Transformers~\cite{parmar2018image,esser2021taming,ding2021cogview,yu2022scaling}, NATs natively support producing images of decent quality with only 4 to 8 steps.
This ability of few-step sampling mainly stems from their ``parallel decoding'' mechanism, which allows multiple tokens to be decoded simultaneously in each step, as shown in Figure~\hyperref[fig:teaser]{\ref*{fig:teaser}a}.
Nevertheless, this mechanism introduces intricate design considerations, such as determining the number and subset of tokens to decode at each step and setting the temperatures for sampling.
Effectively managing these aspects necessitates a complex generation policy comprising numerous hyperparameters, making it difficult for practitioners to utilize these models properly.
Existing works alleviate this problem by \emph{manually} developing multiple scheduling functions for policy configuration.
However, this approach demands expertise and extensive effort, yet still falls short of capturing the optimal dynamics of the generation process (see Table~\ref{tab:effectiveness}).
Furthermore, the \emph{globally shared} policy may lack the flexibility to accommodate the diverse characteristics of different samples.

To address these aforementioned issues, this paper presents \Ours.
The major insight behind \Ours is to consider a \emph{learnable} policy network that automatically configures the policy \emph{adaptively} conditioned on each sample.
Consequently, the generation policy tailored for each sample can be obtained with minimal human effort.
To implement our idea, a non-trivial challenge lies in designing an effective algorithm to train the policy network.
More precisely, due to the non-differentiable nature of the discrete token-based generation process, it is infeasible to straightforwardly adopt standard end-to-end optimization techniques~\cite{ruder2016overview}.
Hence, we propose to formulate the determination of the optimal generation policy as a Markov decision process (MDP).
As a result, our policy network can be naturally defined as an agent, which observes the status of generation, adaptively configures the policy to maximize generation quality, and can be trained through reinforcement learning (\emph{e.g.}, policy gradient).

Importantly, in our problem, designing appropriate reward signals is crucial for training the policy network effectively.
To investigate this issue, we first consider two off-the-shelf design choices: 1) standard evaluation metrics like Fr\'echet Inception Distance (FID)~\cite{heusel2017gans} 2) pre-trained image reward models~\cite{xu2024imagereward}.
We experimentally demonstrate that with these designs, although the expected value of rewards can be successfully maximized, our resulting generative model usually fails to produce sufficiently high-quality and diversified images (see Figure~\ref{fig:comp_rewards}).
In other words, the policy network tends to ``overfit'' these rewards.
Inspired by this phenomenon, we hypothesize that it might be a rescue to consider a reward that is \emph{updated concurrently with the policy network during the learning process} and \emph{adjusted dynamically under the goal of resisting overfitting}.
Therefore, we propose an adversarial reward model, which is formulated as a discriminator similar to that used in generative adversarial networks (GANs)~\cite{goodfellow2014generative}.
When the policy network learns to maximize the reward, we refine the reward model simultaneously to better distinguish between real and generated samples.
In this way, the policy network is effectively prevented from overfitting a static objective, and we observe a balanced diversity and fidelity of the generated images.
An overview of our proposed \Ours is illustrated in Figure~\hyperref[fig:teaser]{\ref*{fig:teaser}c}.

Empirically, the effectiveness of \Ours{} is extensively validated on four benchmark datasets, \emph{i.e.}, ImageNet 256$\times$256~\cite{russakovsky2015imagenet}, ImageNet 512$\times$512~\cite{russakovsky2015imagenet}, MS-COCO~\cite{lin2014microsoft}, and CC3M~\cite{sharma2018conceptual}.
\Ours is able to adaptively adjust the generation policy (see Figure~\ref{fig:vis_ada}) and improve the performance of NATs considerably (\emph{e.g.}, 40\% relative improvement, see Table~\ref{tab:effectiveness}).

\section{Related Work}

\paragraph{Reinforcement learning in image generation.}
The integration of reinforcement learning (RL) for image generation began with early works like~\cite{bachman2015data}. Recently, ImageReward~\cite{xu2024imagereward} collected a large-scale human preference dataset to train a reward model. This pre-trained reward model has spurred research on using RL to fine-tune diffusion-based image generation models~\cite{black2023training,fan2024reinforcement,zhang2024large}. Unlike these methods, which optimize the generative model itself, our work uses an RL agent to enhance a frozen generative model (NAT) without tuning it.

\paragraph{Generative adversarial networks and RL.}
There are several works that use ideas from reinforcement learning to train GANs~\cite{yu2017seqgan,wang2017irgan,wang2018no,sarmad2019rl} for text generation, information retrieval, point cloud completion, \etc.
Recently, SFT-PG~\cite{fan2023optimizing} combines RL with a GAN objective for diffusion-based image generation.Our work differs from these works in many important aspects.
First, the main idea is orthogonal. We are interested in exploring a better generation policy \emph{given} a pre-trained generator backbone (which is kept unchanged throughout our method), while previous studies investigate the alternative ``RL+GAN'' approach to directly train or fine-tune the generator backbone itself.
Second, the research problem is different. We focus on token-based generation, while previous works aim to improve GAN or diffusion-based models.
Third, compared to SFT-PG~\cite{fan2023optimizing}, we conduct comparative analyses on different reward designs and large-scale experiments while they mainly explored the GAN-based objective and prove their concept on relatively simple datasets (\eg, CIFAR~\cite{krizhevsky2009learning} and CelebA~\cite{liu2015deep}).

\paragraph{Non-autoregressive Transformers (NATs)} find their roots in machine translation for their quick inference abilities~\cite{ghazvininejad2019mask, gu2020fully}. These models are recently employed for image synthesis, producing decent-quality images efficiently~\cite{chang2022maskgit,lezama2022improved,li2023mage,chang2023muse,qian2023strait,yu2023language}.
As a pioneering work, MaskGIT~\cite{chang2022maskgit} first demonstrates NAT's effectiveness on ImageNet.
It has been further extended for text-to-image generation and scaled up to 3B parameters in Muse~\cite{chang2023muse} and yields remarkable performance.
MAGE~\cite{li2023mage} proposes leveraging NATs to unify representation learning with image synthesis.
More recently, AutoNAT~\cite{ni2024revisiting} optimizes the policies in NATs with FID~\cite{heusel2017gans} as the objective.
However, as we will show in this paper, the FID-based objective may lead to degenerate solutions.
We present more comparisons with AutoNAT in Appendix~\ref{sec:adanat_vs_autonat}.

\section{Preliminaries of Non-autoregressive Transformers}
\label{sec:preliminaries}
In this section, we briefly introduce the non-autoregressive Transformers (NATs) for image generation, laying the basis for our proposed \Ours.
\subsection{Overview}
Non-autoregressive Transformers (NATs) typically utilize a pre-trained VQ autoencoder~\cite{van2017neural,razavi2019generating,esser2021taming} to convert images into discrete visual tokens and vice versa. The VQ autoencoder comprises an encoder $\mathcal{E}^{\text{VQ}}$, a quantizer $\mathcal{Q}$ with a learnable codebook $e$, and a decoder $\mathcal{D}^{\text{VQ}}$. The encoder and quantizer transform an image into a sequence of visual tokens, while the decoder reconstructs the image from these tokens. NATs generate visual tokens in the latent VQ space by training with the masked token modeling (MLM) objective from BERT~\cite{devlin2019bert}. This allows the model to predict tokens based on surrounding unmasked tokens, enabling multi-step token generation conditioned on previously predicted tokens.

\subsection{Generation via Parallel Decoding}
\label{sec:prelim_parallel_decoding}
During inference, NATs generate the latent visual tokens iteratively.
Specifically, we denote the visual tokens obtained by the VQ encoder as $\bm{v}=[v_i]_{i=1:N}$, where $N$ is the sequence length.
Each visual token $v_i$ corresponds to a specific index of the VQ encoder's codebook.
The model starts from an all-\texttt{[MASK]} token sequence $\bm{v}^{(0)}$.
At $t^{\textnormal{th}}$ step, the model predicts $\bm{v}^{(t+1)}$ from $\bm{v}^{(t)}$ by first \textbf{parallely decoding} all tokens and then \textbf{re-masking} less reliable predictions, as described below\footnote{For the ease of notation, we omit condition $\bm{c}$.}.

\paragraph{Parallel decoding.}  Given visual tokens $\bm{v}^{(t)}$, the model first parallely decodes all of the \texttt{[MASK]} tokens to form an initial guess $\hat{\bm{v}}^{(t+1)}$:
\begin{equation*}
    \hat{v}^{(t+1)}_i
    \begin{cases}
        \sim  \hat {p}_{{\color{sampt}\tau_1^{(t)}}}(v_i|\bm{v}^{{(t)}}),  & \text{if $v_{i}^{(t)} = $\texttt{[MASK]}}; \\
        = v^{(t)}_i,  & \text{otherwise}.
    \end{cases}
\end{equation*}
Here, $\hat {p}_{{\color{sampt}\tau_1^{(t)}}}(v_i|\bm{v}^{{(t)}})$ represents the model's predicted probability distribution at position $i$, scaled by a temperature ${\color{sampt}\tau_1^{(t)}}$.
Meanwhile, confidence scores $\bm{c}^{(t)}$ are defined for all tokens:
\begin{equation*}
{c}_i^{(t)} =
\begin{cases}
    \log \hat{p}(v_i=\hat{v}_i^{(t+1)}|\bm{v}^{(t)}),  & \text{if $v_{i}^{(t)} = $\texttt{[MASK]}};  \\
    +\infty,  & \text{otherwise}.
\end{cases}
\end{equation*}
where $\hat{p}(v_i=\hat{v}_i^{(t+1)}|\bm{v}^{(t)})$ is the predicted probability for the selected token $\hat{v}_i^{(t+1)}$ at position $i$.

\paragraph{Re-masking.}  From the initial guess $\hat{\bm{v}}^{(t+1)}$, the model then obtains $\bm{v}^{(t+1)}$ by re-masking the $\lceil {\color{ratio}m^{(t)}}\cdot N \rceil$ least confident predictions:
\begin{equation*}
    v^{(t+1)}_i =
    \begin{cases}
        \hat{v}^{(t+1)}_i, & \text{if $i \in \mathcal{I}$}; \\
        \texttt{[MASK]} ,  & \text{if $i \notin \mathcal{I}$}.
    \end{cases}
\end{equation*}
Here, \( {\color{ratio}m^{(t)}} \in [0, 1] \) regulates the proportion of tokens to be re-masked at each step. The set \(\mathcal{I}\) comprises indices of the \(N-\lceil {\color{ratio}m^{(t)}}\cdot N \rceil\) most confident predictions and are sampled without replacement from $\text{Softmax}(\bm{c}^{(t)}/\tau_2^{(t)})$\footnote{In practice, this sampling procedure is implemented via Gumbel-Top-$k$ trick~\cite{kool2019stochastic}.}, where $\color{remaskt}\tau_2{(\cdot)}$ is the temperature parameter for re-masking.

The model iterates the process for \( T \) steps to decode all \texttt{[MASK]} tokens, yielding the final sequence \( \bm{v}^{(T)} \).
The sequence is then fed into the VQ decoder to obtain the image $\bm{x}$:
\begin{equation*}
    {\bm{x}} = \mathcal{D}^{\text{VQ}}(\bm{v}^{(T)}).
\end{equation*}

\paragraph{Classifier-free guidance.}  
Notably, NATs can also employ classifier-free guidance (CFG)~\cite{ho2022classifier,chang2023muse} to improve generation quality. This is achieved by extrapolating the logits during the parallel-decoding phase of each timestep $t$ with a guidance scale $\color{cfg}w^{(t)}$.
We refer readers to~\cite{chang2023muse} for more details.

\section{AdaNAT}

\subsection{Motivation}
\begin{align}
    \textcolor{gray}{\textsf{Existing works}:} \quad & \textcolor{gray}{m^{(t)}, \tau_1^{(t)}, \tau_2^{(t)}, w^{(t)} = \bm{\eta}(t)} \quad & \textcolor{gray}{\textsf{\textit{pre-defined}, \textit{static}}} \label{eq:exist}\\
    \text{\textbf{\textsf{AdaNAT}}:} \quad & m^{(t)}, \tau_1^{(t)}, \tau_2^{(t)}, w^{(t)} = \bm{\eta}_{\textcolor{learnable}{{\bm{\phi}}}}(t, \textcolor{adaptive}{\bm{v}^{(t)}}) \quad & \text{\learnable, \adaptive} \label{eq:optimal}
\end{align}
\textbf{Motivation I: automatic policy acquisition.}
The flexible parallel decoding scheme endows NATs with an inherent ability for efficient image generation.
However, it necessitates an intricate generation policy comprising a variety of hyperparameters for meticulous control.
As discussed in Section~\ref{sec:preliminaries}, a $T$-step generation process introduces $4\times T$ hyperparameters: $\{m^{(t)}, \tau_1^{(t)}, \tau_2^{(t)}, w^{(t)}\}_{t=0}^{T-1}$.
The abundance of these hyperparameters makes it difficult for practitioners to take full advantage of NATs in realistic scenarios.
Existing works mainly alleviate this problem by leveraging multiple \emph{pre-defined} scheduling functions (denoted together as $\bm{\eta}$) to manually configure these hyperparameters (Eq.~\ref{eq:exist}, detailed in Appendix~\ref{sec:scheduling}).
Nevertheless, such a manual-design regime relies on a fair amount of expert knowledge and labor efforts, yet still leads to a notably sub-optimal policy (see Table~\ref{tab:effectiveness}).
To address this issue, we propose to consider a learnable policy net $\bm{\eta}_{\bm{\phi}}$ that is trained to produce the appropriate policy automatically.
As a result, a significantly superior generation policy can be acquired with minimal human efforts once a proper learning algorithm for $\bm{\eta}_{\bm{\phi}}$ is utilized (which will be discussed later).

\paragraph{Motivation II: adaptive policy adjustment.}
Furthermore, it is noteworthy that Eq.~\ref{eq:exist} is a \emph{static} formulation, \ie, the generation of all samples shares the same groups of scheduling functions.
In contrast, we argue that every sample has its own characteristics, and ideally the generation process should be adaptively adjusted according to each individual sample.
To attain this goal, we propose to dynamically determine the policy for $t^{\textnormal{th}}$ generation step conditioned on the current generation status, \ie, the current visual tokens $\bm{v}^{(t)}$.
In other words, $\bm{v}^{(t)}$ provides necessary information on how the generated sample `looks like' at $t^{\textnormal{th}}$ step, based on which a tailored policy can be derived to better enhance the generation quality.
More evidence to validate the rationality of our idea can be found in Table~\ref{tab:effectiveness} and Figure~\ref{fig:vis_ada}.

Integrating the discussions above, we propose to establish a policy network $\bm{\eta}_{\bm{\phi}}$ that directly \emph{learns} to produce the appropriate policy in a \emph{adaptive} manner, as shown in Eq.~\ref{eq:optimal}. In the following, we will introduce how to train $\bm{\eta}_{\bm{\phi}}$ effectively.

\subsection{Policy Network Optimization}

Formally, given a pre-trained NAT model parameterized by $\bm{\theta}$ and a policy network $\bm{\eta}_{\bm{\phi}}$ to be trained, our objective is to maximize the expected quality of the generated images:
\begin{equation}
    \underset{{\bm{\phi}}}{\text{maximize}} \quad J({\bm{\phi}}) = \mathbb{E}_{\bm{x}\sim p_{\bm{\theta}}(\bm{x}\mid \bm{\eta}_{{\bm{\phi}}})} [r(\bm{x})],\label{eq:policy overall objective init}
\end{equation}
where $r(\cdot)$ is a function quantifying image quality, as detailed in Section~\ref{sec:reward}, and $p_{\bm{\theta}}(\bm{x}\mid \bm{\eta}_{{\bm{\phi}}})$ denotes the distribution of the images generated by the NAT model $\bm{\theta}$, under the generation policy specified by the policy network $\bm{\eta}_{\bm{\phi}}$.

It is challenging to solve problem \ref{eq:policy overall objective init} directly.
This is caused by the non-differentiability of the generation process, attributed to the discrete nature of the visual tokens in NATs.
To address this issue, we propose to reformulate the determination of the generation policy as a Markov decision process (MDP).
Under this framework, the policy network $\bm{\eta}_{{\bm{\phi}}}$ can be naturally defined as an agent that observes the generation status and takes actions to configure the policy, which can then be optimized via reinforcement learning.

\paragraph{Markov Decision Process (MDP) formulation.}
We model the NATs' generation process as a $T$-horizon Markov Decision Process: $MDP(\mathcal{S}, \mathcal{A}, P, R)$, where:
\begin{itemize}[topsep=0pt]
    \item $\mathcal{S}$ denotes the state space. Each state $\bm{s}_t \in \mathcal{S}$ is defined as:
    \begin{equation}
        \bm{s}_t \triangleq (t, \bm{v}^{(t)}),
    \end{equation}
    where $\bm{v}^{(t)}$ is the visual tokens at time $t$.
    \item $\mathcal{A}$ represents the action space, with each action $\bm{a}_t \in \mathcal{A}$ corresponding to the generation policy at time $t$:
    \begin{equation}
        \bm{a}_t \triangleq  \big(m^{(t)}, \tau_1^{(t)}, \tau_2^{(t)}, w^{(t)}\big)
    \end{equation}
    \item $P: \mathcal{S} \times \mathcal{A} \times \mathcal{S} \rightarrow [0, 1]$ is the state transition probability function. Given the current state $\bm{s}_t$ and action $\bm{a}_t$, $P$ models the probability of transitioning to the next state $\bm{s}_{t+1}$:
    \begin{equation}
        P(\bm{s}_{t+1} \mid \bm{s}_t, \bm{a}_t) \triangleq \big (\delta_{t+1}, \mathcal{T}(\bm{v}^{(t+1)} | \bm{v}^{(t)}; \bm{a}_t, \bm{\theta})\big),
    \end{equation}
    where $\delta_{t+1}(\cdot)$ is the Dirac delta function that is nonzero only at $t+1$. $\mathcal{T}(\bm{v}^{(t+1)} | \bm{v}^{(t)}; \bm{a}_t, \bm{\theta})\big)$ denotes the transition distribution from visual tokens $\bm{v}^{(t)}$ to $\bm{v}^{(t+1)}$. This term encapsulates the visual token transition process as detailed in Section~\ref{sec:prelim_parallel_decoding}.
    \item $R: \mathcal{S} \times \mathcal{A} \rightarrow \mathbb{R}$ is the reward function, designed to reflect the quality of the visual output. The reward function is defined as:
    \begin{equation}
        R(\bm{s}_t, \bm{a}_t) \triangleq r\big(\mathcal{D}^{\text{VQ}}(\bm{v}^{(t)})\big) \cdot \mathbb{I}_{\{t = T\}},
    \end{equation}
    where $r(\cdot)$ measures the quality of the generated image and $\mathcal{D}^{\text{VQ}}$ is the VQ-decoder that converts visual tokens to pixels. This formulation ensures that the reward is assessed only at the final timestep $T$, as we are only concerned with the quality of the final output.
    We describe the reward function design in more detail in Section~\ref{sec:reward}.
\end{itemize}

\paragraph{Learning the adaptive policy.}
Given the MDP formulation, we are able to formalize the policy network as an agent $\bm{\pi}_{{\bm{\phi}}}$ that takes actions to configure the generation policy based on the current state $\bm{s}_t$:
\begin{equation}
    m^{(t)}, \tau_1^{(t)}, \tau_2^{(t)}, w^{(t)} \sim \bm{\pi}_{\bm{\phi}}(\bm{a}_t \mid \bm{s}_t)
\end{equation}
As $\bm{s}_t$ includes both the decoding step $t$ and the current visual tokens $\bm{v}^{(t)}$, the policy network is able to perform adaptive configuration selection.
Notably, the agent is defined as a stochastic version of the original policy network $\bm{\eta}_{{\bm{\phi}}}$ to balance exploration and exploitation in reinforcement learning:
\begin{equation}
    \bm{\pi}_{\bm{\phi}}(\bm{a}_t \mid \bm{s}_t) \triangleq \mathcal{N}\big(\bm{\eta}_{\bm{\phi}}(\bm{s}_t), \sigma \bm{I}\big),
\end{equation}
where $\mathcal{N}(\cdot)$ denotes a multivariate normal distribution, and $\sigma$ is a hyperparameter controlling the exploration level.
At test time, we simply take the mean of the distribution as the action $\bm{a}_t$, which reduces to the deterministic policy $\bm{\eta}_{\bm{\phi}}(\bm{s}_t)$.

We train $\bm{\pi}_{\bm{\phi}}$ to maximize the expected reward over the generation process:
\begin{equation}
J({\bm{\phi}}) = \mathbb{E}_{\bm{\pi}_{\bm{\phi}}}\left[ \sum_{t=0}^{T} R(\bm{s}_t, \bm{a}_t) \right]= \mathbb{E}_{\bm{\pi}_{\bm{\phi}}}\big[R(\bm{s}_T, \bm{a}_T)\big],
\end{equation}
To effectively estimate policy gradient, we generally follow the clipped surrogate objective in Proximal Policy Optimization (PPO) algorithm~\cite{schulman2017proximal}:
\begin{equation}
    \begin{split}
    L^{PPO}({\bm{\phi}}) = \mathbb{E}_{t} \Bigg[ & \min\left( \rho_t({\bm{\phi}}) \hat{A}_t, \text{clip}\left(\rho_t({\bm{\phi}}), 1-\epsilon, 1+\epsilon\right) \hat{A}_t \right) \\
    & - c \big( V_{{\bm{\phi}}}(\bm{s}_t) - R(\bm{s}_T, \bm{a}_T) \big)^2 \Bigg],
    \end{split}
    \label{eq:ppo}
\end{equation}
where $\rho_t({\bm{\phi}}) = \frac{\bm{\pi}_{\bm{\phi}}(\bm{a}_t \mid \bm{s}_t)}{\bm{\pi}_{{\bm{\phi}}_{\text{old}}}(\bm{a}_t \mid \bm{s}_t)}$ denotes the probability ratio of the new policy to the old policy for taking action $\bm{a}_t$ in state $\bm{s}_t$, $V(\bm{s}_t)$ is a learned state-value function, $\hat{A}_t$ is the advantage estimate at timestep $t$, and $\epsilon, c$ are hyperparameters.
The advantage estimate $\hat{A}_t$ is calculated as:
\begin{equation}
    \hat{A}_t = -V(\bm{s}_t) + R(\bm{s}_T, \bm{a}_T),
\end{equation}
We provide more details about the PPO algorithm in the Appendix~\ref{sec:imp_details}.

\subsection{Reward Design}
\label{sec:reward}
One of the key aspects in training our reinforcement learning agent is the design of the reward function.
In this section, we start with straightforward reward designs and discuss their challenges.
Then, we propose an adversarial reward design that effectively addresses these challenges.

\hypertarget{opt:1}{}
\paragraph{A: Pre-defined evaluation metric.}
The most straightforward reward design is to employ commonly used evaluation metrics in the image generation task such as Fr\'echet Inception Distance (FID).
However, we observe two challenges in practice.
\emph{First}, statistical metrics face challenges in providing sample-wise reward signals.
Evaluation metrics in image generation tasks are usually statistical, \ie, computed over a large number of generated images.
For example, the common practice in ImageNet 256$\times$256 benchmark is to evaluate the FID and IS score over 50K generated images~\cite{peebles2023scalable,bao2022all}, which makes it challenging to attribute the reward to specific actions.
In practice, we find this lack of informative feedback leads to failure in training the adaptive policy network, as detailed in Appendix~\ref{sec:fid_reward}.
\emph{Second}, better metric scores do not necessarily translate into better visual quality.
As shown in Figure~\hyperref[fig:comp_rewards]{\ref*{fig:comp_rewards}a}, even when the evaluation metric FID is successfully optimized (\eg, using a non-adaptive variant of \Ours), the generated images may still suffer from poor visual quality.
This underscores the limitations of directly adopting evaluation metrics as optimization objectives and motivates us to find better alternatives for the reward function.

\hypertarget{opt:2}{}
\paragraph{B: Pre-trained reward model.}
Another option is to adopt a pre-trained, off-the-shelf reward model~\cite{xu2024imagereward} specialized at assessing the visual quality of images.
This approach addresses the two challenges in option \hyperlink{opt:1}{A} as the reward signal is now 1) provided for individual images and 2) more consistent with the image quality.
However, we observe that the generated images in this case tend to converge on a similar style with relatively low diversity, as shown in Figure~\hyperref[fig:comp_rewards]{\ref*{fig:comp_rewards}b}.
Furthermore, establishing a proper image reward model often involves collecting large-scale annotated human preference data and training additional deep networks, a procedure that is generally costly.
Therefore, the access to a pre-trained image reward model may not be assumed in all the scenarios.

\paragraph{Ours: Adversarial reward modeling.}
One shared issue with the aforementioned two designs is that, while the expected value of rewards can be effectively maximized, the resultant images exhibit unintended inferior quality or limited diversity.
In other words, the policy network tends to ``overfit'' these rewards.
Motivated by this observation, we hypothesize that it might be a rescue to consider a reward that is updated concurrently with the policy network during the learning process and adjusted dynamically with the goal of resisting overfitting.
To this end, we propose adversarial reward modeling, where we learn an adversarial reward model $r_{\bm{\psi}}$ together with the policy network.
Specifically, we formulate $r_{\bm{\psi}}$ as a discriminator akin to that in GANs~\cite{goodfellow2014generative} and establish a minimax game between the policy network and $r_{\bm{\psi}}$:
\begin{align}
    \underset{{\bm{\phi}}}{\text{maximize}}\quad J({\bm{\phi}})&=\mathbb{E}_{\bm{x}\sim p_{\bm{\theta}}(\bm{x}\mid \bm{\pi}_{{\bm{\phi}}})} \left[ r_{\bm{\psi}}(\bm{x}) \right],\label{eq:policy overall objective} \quad \text{{\textcolor{gray}{(Maximize Eq.~\ref{eq:ppo} in practice)}}}\\
    \underset{{\bm{\psi}}}{\text{minimize}}\quad L({\bm{\psi}})&=\mathbb{E}_{\bm{x}\sim p_{\bm{\theta}}(\bm{x}\mid \bm{\pi}_{{\bm{\phi}}})} \left[ \log r_{\bm{\psi}}(\bm{x}) \right] + \mathbb{E}_{\bm{x}\sim p_{\text{real}}(\bm{x})} \left[ \log (1 - r_{\bm{\psi}}(\bm{x})) \right],\label{eq:adversarial_reward}
\end{align}
where $p_{\text{real}}$ denotes the distribution of real images.
As the policy tries to maximize the reward, the reward model is refined simultaneously to better distinguish between real and generated samples.
Consequently, we effectively curb the policy network from overfitting a static objective, and more balanced diversity and fidelity of the generated images are also observed (see Figure~\hyperref[fig:comp_rewards]{\ref*{fig:comp_rewards}c}).
Moreover, the adversarial reward model offers immediate, sample-wise reward signals, contrasting with the statistical metrics used in option \hyperlink{opt:1}{A}; and it obviates the need for expensive human preference data required by the pre-training of reward models in option \hyperlink{opt:2}{B}.
We summarize the training procedure for \Ours in Algorithm~\ref{algo:adanat}.

\begin{center}
    \resizebox{.8\columnwidth}{!}{
        \begin{minipage}{.8\linewidth}             \begin{algorithm}[H]
                \caption{Training Procedure for AdaNAT \label{algo:adanat}}
                \begin{algorithmic}[1]
                    \State {\bfseries Input:} Policy network $\bm{\pi}_{\bm{\phi}}$ and adversarial reward model $r_{\bm{\psi}}$
                    \For{$i = 0, 1, 2, \dotsc$}
                        \State \textcolor{blue}{\texttt{\# Policy network optimization}}
                        \State Sample trajectories $\tau\sim\bm{\pi}_{{\bm{\phi}}_\text{old}}$
                        \State Update $\bm{\pi}_{\bm{\phi}}$ with the gradient $\nabla_{\bm{\phi}} L^{PPO}({\bm{\phi}})$ \Comment{Refer to Eq.~\ref{eq:ppo}}
                        \State ${\bm{\phi}}_{\text{old}}$ $\leftarrow$ ${\bm{\phi}}$
                        \State \textcolor{blue}{\texttt{\# Reward model optimization}}
                        \State Sample images $\bm{x}_{\text{real}} \sim p_{\text{real}}$, $\bm{x}_{\text{fake}} \sim p_{\bm{\theta}}(\bm{x} \mid \bm{\pi}_{\bm{\phi}})$
                        \State Update $r_{\bm{\psi}}$ with the gradient $\nabla_{\bm{\psi}} L({\bm{\psi}})$ \Comment{Refer to Eq.~\ref{eq:adversarial_reward}}
                    \EndFor
                \end{algorithmic}
            \end{algorithm}
        \end{minipage}
    }
\end{center}

\section{Experiments}

\paragraph{Setups.}
Consistent with prior work~\cite{chang2022maskgit,li2023mage,chang2023muse}, we utilize a pretrained VQGAN~\cite{esser2021taming} with a 1024-codebook for image-token conversion.
Our NAT models adopt the U-ViT~\cite{bao2022all} architecture, a Transformer type tailored for image generation, in two sizes: \Ours-S (13 layers, 512 dimensions) and \Ours-L (25 layers, 768 dimensions).
We use a patch size of 2 for ImageNet 512$\times$512 to manage the higher token count.
We perform the optimization loop of \Ours in Algorithm~\ref{algo:adanat} for 1000 iterations using Adam optimizer~\cite{kingma2014adam}.
In practice, our adaptive policy network reuses the off-the-shelf NAT model's output feature $f_{\bm{\theta}}(\bm{v}^{(t)})$ as the generation status inputs, which we find to be more efficient.
Notably, the pre-trained NAT model is kept fixed and there is no need to back-propagate gradient through it throughout the optimization process.
For class-conditional generation on ImageNet~\cite{russakovsky2015imagenet}, we report FID-50K following~\cite{dhariwal2021diffusion,peebles2023scalable}.
For text-to-image generation on MS-COCO~\cite{lin2014microsoft} and CC3M~\cite{sharma2018conceptual}, we report FID-30K following~\cite{bao2022all,chang2023muse}.
Due to space limitation, the specifics of NATs pre-training and more implementation details of our method are deferred in Appendix~\ref{sec:imp_details}.

\subsection{Main Results}
\begin{table*}[!t]
    \caption{
        \textbf{Class-conditional image generation on ImageNet 256$\times$256 \label{tab:fid_imnet}}.
        TFLOPs quantify the computational cost for generating a single image.
        For DPM-Solver~\cite{lu2022dpm} augmented diffusion models (marked with $^\dagger$), we follow~\cite{lu2022dpm} to tune configurations and report the lowest FID.
        $^\ddagger$: methods without classifier-based or classifier-free guidance~\cite{dhariwal2021diffusion,ho2022classifier}.
        Diff: diffusion, AR: autoregressive, NAT: non-autoregressive Transformers.
    }
    \centering
    \tablestyle{5pt}{1.05}
    \resizebox{.98\columnwidth}{!}{
        \begin{tabular}{y{140}x{30}x{46}x{27}x{45}x{41}x{20}}
            \toprule
            Method & Type &  Params & Steps & TFLOPs$\downarrow$ & FID-50K$\downarrow$ & IS$\uparrow$  \\\midrule
            \VQVAE             & AR    & 13.5B & 5120    & -    & 31.1  & $\sim$ 45    \\
            \VQGAN             & AR    & 1.4B  & 256     & -    & 15.78 & 78.3    \\
            ADM-G~\cite{dhariwal2021diffusion}\pub{NeurIPS'21}               & Diff. & 554M  & 250     & 334.0  & 4.59  & 186.7  \\
            ADM-G, ADM-U~\cite{dhariwal2021diffusion}\pub{NeurIPS'21} & Diff. & 608M  & 250 & 239.5    & 3.94  & 215.8    \\
            \LDM               & Diff. & 400M  & 250     & 52.3 & 3.60  & 247.7      \\
            \VQDiffusion       & Diff. & 554M  & 100     & 12.4 & 11.89  & -          \\
            \DraftRevise       & NAT   & 1.4B  & 72      & -    & 3.41  & 224.6     \\\hline
            \multirow{2}{*}{\ADMp} & \multirow{2}{*}{Diff.} & \multirow{2}{*}{554M} & 4 & 5.3 & 22.35 & -  \\
            & & & 8 & 10.7 & 8.81 & 174.2 \\
            \multirow{2}{*}{\LDMp} & \multirow{2}{*}{Diff.} & \multirow{2}{*}{400M} & 4 & 1.2 & 11.74 & - \\
            & & & 8 & 2.0 & 4.56 & 262.9   \\
            \multirow{2}{*}{\UVITp} & \multirow{2}{*}{Diff.} & \multirow{2}{*}{501M} & 4 & 1.4 & 8.45 & - \\
            & & & 8 & 2.4 & 3.37 & 235.9 \\
            \multirow{2}{*}{\DITp} & \multirow{2}{*}{Diff.} & \multirow{2}{*}{675M} & 4 & 1.3 & 9.71 & -  \\
            
            & & & 8 & 2.2 & 5.18 & 213.0  \\
            \USF & Diff. & 554M & 8 & 10.7 & 9.72 & - \\
            \MaskGIT & NAT & 227M & 8 & 0.6 & 6.18 & 182.1  \\
            \MaskgitCfg  & NAT & 227M & 12 & 1.4 & 4.92 & - \\
            {MaskGIT-RS~\cite{chang2022maskgit}\pub{CVPR'22}} & {NAT} & {227M} & {8} & {8.4} & {4.02} & {-} \\
                                    \TokenCritic & NAT & 422M & 36 & 1.9 & 4.69 & 174.5 \\
            {Token-Critic-RS~\cite{lezama2022improved}\pub{ECCV'22}} & {NAT} & {422M} & {36} & {8.7} & {3.75} & {-}  \\
            \MAGE & NAT & 230M & 20 & 1.0 & 6.93 & -   \\
            \MaskgitFSQ & NAT & 225M & 12 & 0.8 & 4.53 & - \\\hline
                        \multirow{2}{*}{\textbf{\Ours-S}} & \multirow{2}{*}{NAT} & \multirow{2}{*}{58M} & 4 & \textbf{0.2} & 4.54 & - \\
            & & & 8 & 0.3 & 3.71 & 224.5 \\
            \multirow{2}{*}{\textbf{\Ours-L}} & \multirow{2}{*}{NAT} & \multirow{2}{*}{206M} & 4 & 0.5 & 3.63 & - \\
            & & & 8 & 0.9 & \textbf{2.86} & 265.4 \\\bottomrule
        \end{tabular}
    }
\end{table*}

\paragraph{Class-conditional generation on ImageNet.}
In Tables~\ref{tab:fid_imnet} and \ref{tab:fid_imnet512}, we compare our approach with other generative models on ImageNet 256$\times$256 and 512$\times$512, respectively. Despite having fewer parameters and lower inference costs, our \Ours-S achieves a competitive FID of 4.54 on ImageNet 256$\times$256. With more computational budget, \eg, 0.3 TFLOPs, \Ours-S improves its FID to 3.71, outperforming most baselines. The \Ours-L model furthers this improvement, reaching an FID of 2.86 with 8 steps. On ImageNet 512$\times$512, our top-performing model secures an FID of 3.66, surpassing leading models while requiring substantially less computational effort.
We also compare the practical latency between \Ours and several competitive baselines in Appendix~\ref{sec:practical_latency}.

\paragraph{Text-to-image generation on MS-COCO and CC3M.}
The efficacy of \Ours in text-to-image generation is demonstrated on both MS-COCO~\cite{lin2014microsoft} and CC3M~\cite{sharma2018conceptual}. On MS-COCO, \Ours-S achieves a FID score of 5.75 with only 0.3 TFLOPs, surpassing competing baselines and even outperforming recent diffusion models~\cite{bao2022all, lu2022dpmp} with lower computational resources. On CC3M, \Ours outperforms the advanced non-autoregressive Transformer Muse~\cite{chang2023muse}, achieving a FID of 6.83 versus 7.67, and maintains superior performance even when the computational budget of the Muse model is doubled.

\begin{table*}[!t]
    \caption{
        \textbf{Class-conditional image generation on ImageNet 512$\times$512\label{tab:fid_imnet512}}.
        $^{\dagger}$: DPM-Solver~\cite{lu2022dpm} augmented diffusion models.
        $^\ddagger$: methods without classifier-based or classifier-free guidance~\cite{dhariwal2021diffusion,ho2022classifier}.
        See Table~\ref{tab:fid_imnet} for more details.
    }
    \centering
    \tablestyle{5pt}{1.1}
    \resizebox{.93\columnwidth}{!}{
        \begin{tabular}{y{140}x{30}x{46}x{27}x{45}x{41}x{20}}
            \toprule
            Method & Type &  Params & Steps & TFLOPs$\downarrow$ & FID-50K$\downarrow$ & IS$\uparrow$  \\\midrule
            \VQGAN             & AR    & 227M & 1024    & -     & 26.52 & 66.8   \\
            ADM-G~\cite{dhariwal2021diffusion}\pub{NeurIPS'21}               & Diff.  & 559M & 250     & 579.0 & 7.72  & 172.7 \\
            ADM-G, ADM-U~\cite{dhariwal2021diffusion}\pub{NeurIPS'21} & Diff.   & 731M & 250 & 719.0 & 3.85  & 221.7  \\\hline
            \ADMp & Diff. &  559M 
                          & 8 & 18.5 & 16.16 & 109.2 \\
            
            \UVITp & Diff. &  501M
                        & 8 & 3.4 & 4.60 & 286.8 \\
            \DITp & Diff. &  675M 
                        & 8 & 9.6 & 5.44 & 275.0 \\
            \MaskGIT & NAT  & 227M & 12 & 3.3 & 7.32 & 156.0 \\
            {MaskGIT-RS~\cite{chang2022maskgit}\pub{CVPR'22}} & {NAT} & {227M} & {12} & {13.1} & {4.46} & {-} \\
            \TokenCritic & NAT & 422M & 36 & 7.6 & 6.80 & 182.1  \\
            {Token-Critic-RS~\cite{lezama2022improved}\pub{ECCV'22}} & {NAT} &  {422M} & {36} & {34.8} & {4.03} & {-}  \\\hline
                        \textbf{\Ours-L} & NAT & 232M & 8 & \textbf{1.2} & \textbf{3.66} & 297.3 \\\bottomrule
        \end{tabular}
    }
\end{table*}

\begin{table}[t!]
\begin{minipage}[t]{0.48\textwidth}
\centering
    \caption{
        \textbf{Text-to-image generation on MS-COCO};
        $^{\dagger}$: DPM-Solver~\cite{lu2022dpm} augmented diffusion models.
    }
    \resizebox{1\columnwidth}{!}{
    \tablestyle{3pt}{1.1}
    \begin{tabular}{y{70}x{29}x{19}x{38}x{41}}
        \toprule
        Method & Params &  Steps & TFLOPs$\downarrow$ & FID-30K$\downarrow$\\\midrule
        VQ-Diffusion~\cite{gu2022vector} & 370M & 100 & - & 13.86 \\
        Frido~\cite{fan2023frido} &512M & 200 & - & 8.97 \\
        U-Net$^\dagger$~\cite{bao2022all} &53M & 50 & - & 7.32 \\
                        \multirow{2}{*}{U-ViT$^\dagger$~\cite{bao2022all}} & \multirow{2}{*}{58M} & 4 & 0.4 & 11.88 \\
         &  & 8 & 0.6 & 6.37 \\\hline
                \textbf{\Ours-S} & 57M & 8 &  \textbf{0.3} & \textbf{5.75} \\\bottomrule
    \end{tabular}
    }
\end{minipage}
\begin{minipage}[t]{.505\textwidth}
    \centering
    \caption{
        \textbf{Text-to-image generation on CC3M}; all models are trained and evaluated on CC3M.\label{tab:fid_cc3m}
    }
    \resizebox{1\columnwidth}{!}{
        \tablestyle{3pt}{1.1}
        \begin{tabular}{y{83}x{29}x{19}x{38}x{41}}
            \toprule
            Method                    & Params             & Steps & TFLOPs$\downarrow$ & FID-30K$\downarrow$   \\\midrule
            VQGAN~\cite{esser2021taming}                     & 600M                  & 256                &   -    & 28.86 \\
            LDM-4~\cite{rombach2022high}                     & 645M                  & 50                  &   -    & 17.01 \\
            Draft-and-revise~\cite{lee2022draft}          & 654M                  & 72                  &   -    & 9.65  \\
                                    \multirow{2}{*}{Muse~\cite{chang2023muse}} & \multirow{2}{*}{500M} & 8                  &    2.8    & 7.67  \\
            &                       & 16                 &    5.4    & 7.01  \\\hline
            \textbf{\Ours-Muse}             & 512M          & 8                  &    2.8    & \textbf{6.83}  \\\bottomrule
        \end{tabular}
    }
\end{minipage}
\end{table}
\subsection{Analysis of \Ours}

This section presents more analyses of \Ours , including its effectiveness, the learned adaptive policy, and comparisons of different reward designs.

\begin{figure*}[t!]\centering
\includegraphics[width=.75\linewidth]{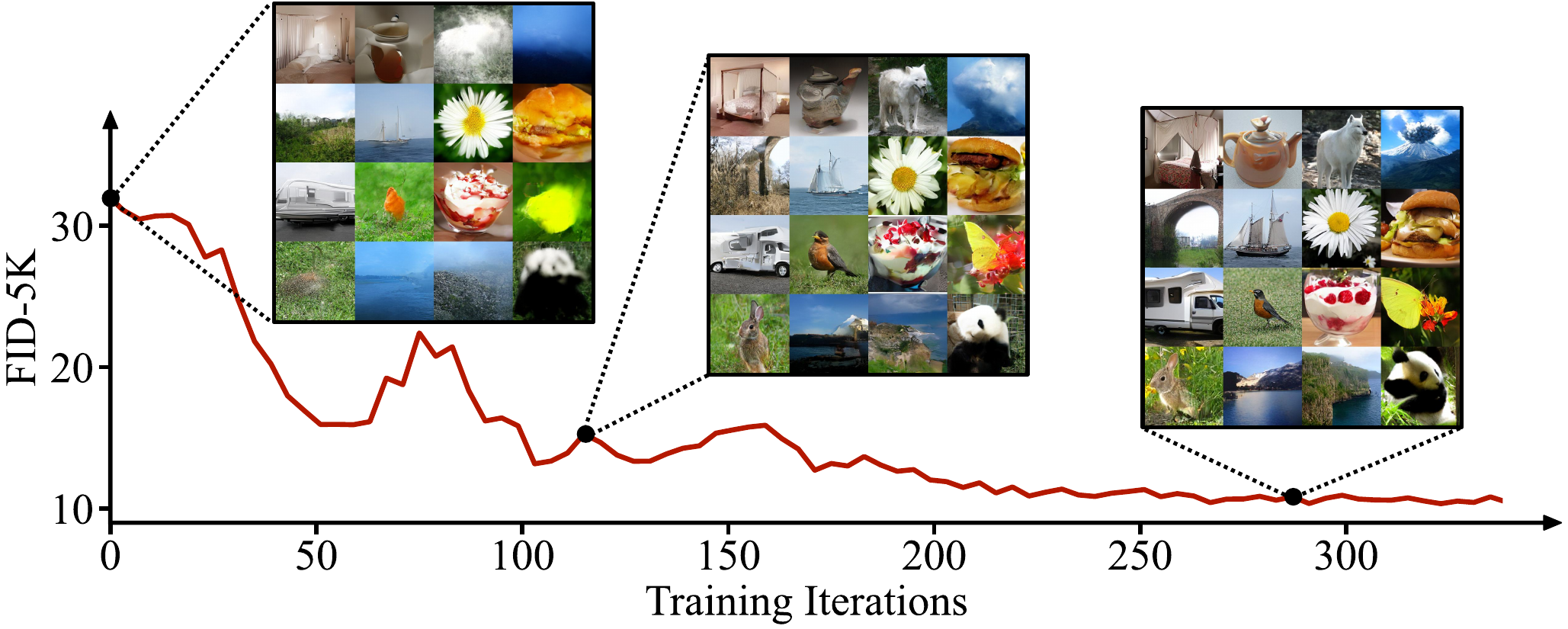}
\caption{
    \textbf{Optimization process of \Ours.}
    We plot the training curve of \Ours-L ($T=4$) on ImageNet 256$\times$256 and visualize samples from different stages.
    We train the policy network to output suitable configuration, while keeping the pre-trained NAT model \emph{fixed} and only use it for inference.
    FID-5K is used for efficient evaluation.
}
\label{fig:optimization_curve}
\end{figure*}
\begin{table}[t]
\centering
\caption{
    \textbf{Effectiveness of \Ours}.
    We use \Ours-S ($T=4$) on ImageNet-256.
    \label{tab:effectiveness}
}
\resizebox{.5\columnwidth}{!}{
    \tablestyle{6pt}{1.15}
    \begin{tabular}{x{50}x{50}|x{50}}
        \toprule
        \multicolumn{2}{c|}{Generation Policy} & \multirow{2}{*}{FID-50K$\downarrow$}\\
        \emph{Learnable?} & \emph{Adaptive?} &  \\\midrule
        \gc{\xmark} & \gc{\xmark} & \gc{7.65} \\
        \cmark     & \xmark & 5.40\fontsize{5.5pt}{0.1em}\selectfont~\hl{(-2.25)} \\
        \cmark & \cmark & \textbf{4.54}\fontsize{5.5pt}{0.1em}\selectfont~\hl{(-3.11)} \\\bottomrule
    \end{tabular}
}
\end{table}

\begin{figure*}[t!]\centering
\includegraphics[width=.8\linewidth]{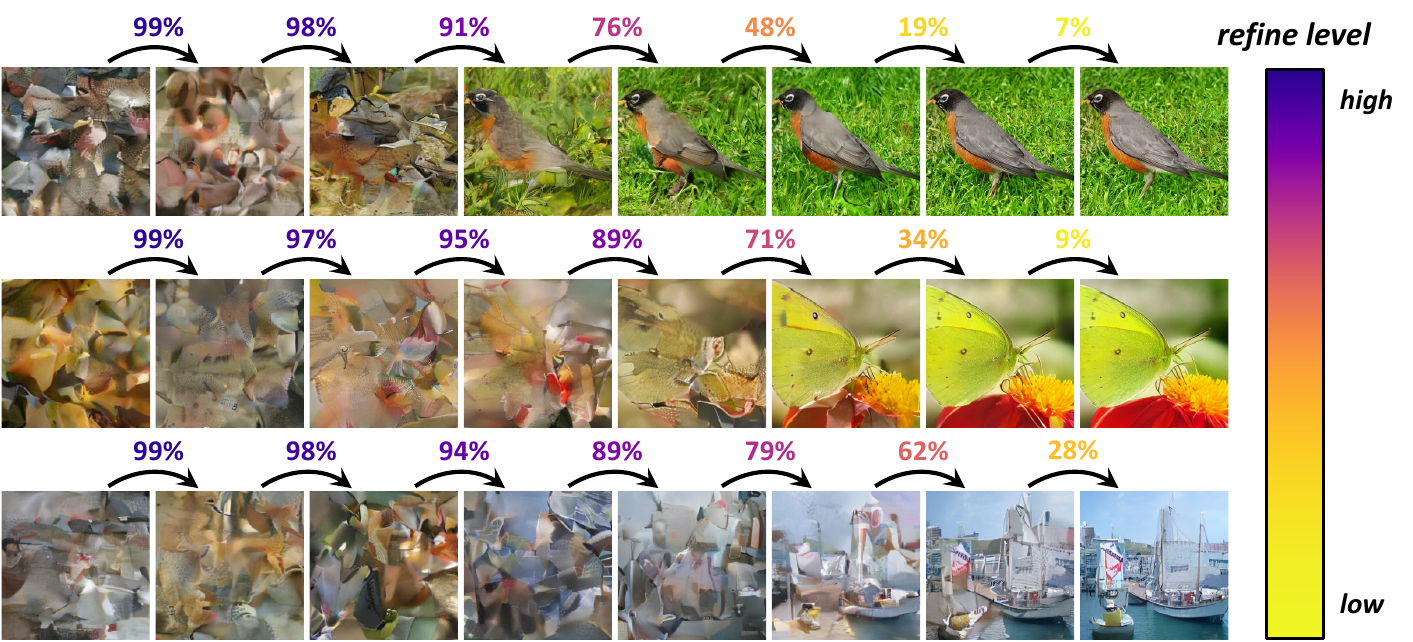}
\caption{
\textbf{Visualizing the adaptive policy.}
The re-masking ratio $m^{(t)}$ (\textbf{refine level}), which controls the proportion of least-confident tokens to be refined at each step, is visualized as an example (see Section~\ref{sec:prelim_parallel_decoding} for $m^{(t)}$'s definition).
The policy network adaptively reduces $m^{(t)}$ for only minor refinements when the sample already reaches a decent quality; otherwise, it keeps adopting relatively higher $m^{(t)}$ for more adjustments.
}
\label{fig:vis_ada}
\end{figure*}

\paragraph{Effectiveness of \Ours.}
\label{sec:effectiveness}
In Table~\ref{tab:effectiveness}, we analyze the effectiveness of \Ours.
Incorporating learnability into the generation policy alone leads to a 2.25 decrease in FID, marking a 30\% improvement over the baseline.
This highlights the potential sub-optimality of the manual designs in prior works~\cite{chang2022maskgit,lezama2022improved,li2023mage,chang2023muse}.
When the adaptive mechanism is introduced, the FID score is further reduced to 4.54, achieving an additional 16\% improvement over the learnable setup.
The above results underscore the effectiveness of a learnable, adaptive policy in the generation process.
We offer additional qualitative results in Figure~\ref{fig:vis_ada} to further illustrate this point.

In Figure~\ref{fig:optimization_curve}, we present the optimization curve of \Ours, together with images sampled at different stages of the training process.
Initially, the policy network's policy lead to very blurry images, with high FID scores.
As training progresses, the policy network gradually refines itself, resulting in lower FID scores and a perceptible improvement in image quality, despite occasional distortions.
Finally, the policy network converges with a robust generation policy that yields low FID scores and produces images of decent quality consistently.
This offers a more comprehensive understanding of \Ours's efficacy in navigating towards more feasible generation policy for non-autoregressive Transformers.

\paragraph{Visualizing the learned adaptive policy.}
In Figure~\ref{fig:vis_ada}, we visualize the learned adaptive policy, taking the re-masking ratio $m^{(t)}$ as a representative example.
The re-masking ratio controls the proportion of least-confident tokens to be refined at each step, so we also call it ``\textbf{refine level}''.
For the sample with a relatively simple structure (first row), the refine level is rapidly reduced to a lower level as the sample reaches a decent quality, which restricts the NAT model to only make minor refinements.
When the structure of the sample becomes harder (second row), more steps are required to reach a satisfactory quality, and the policy network keeps adopting relatively higher refine levels for more adjustments.
For a sample with a complex structure (third row), the policy network consistently applies higher refine levels to make more adjustments till the end.
These visualizations provide deeper insights into the adaptive mechanism of \Ours.

\begin{figure*}[t!]\centering
\includegraphics[width=\linewidth]{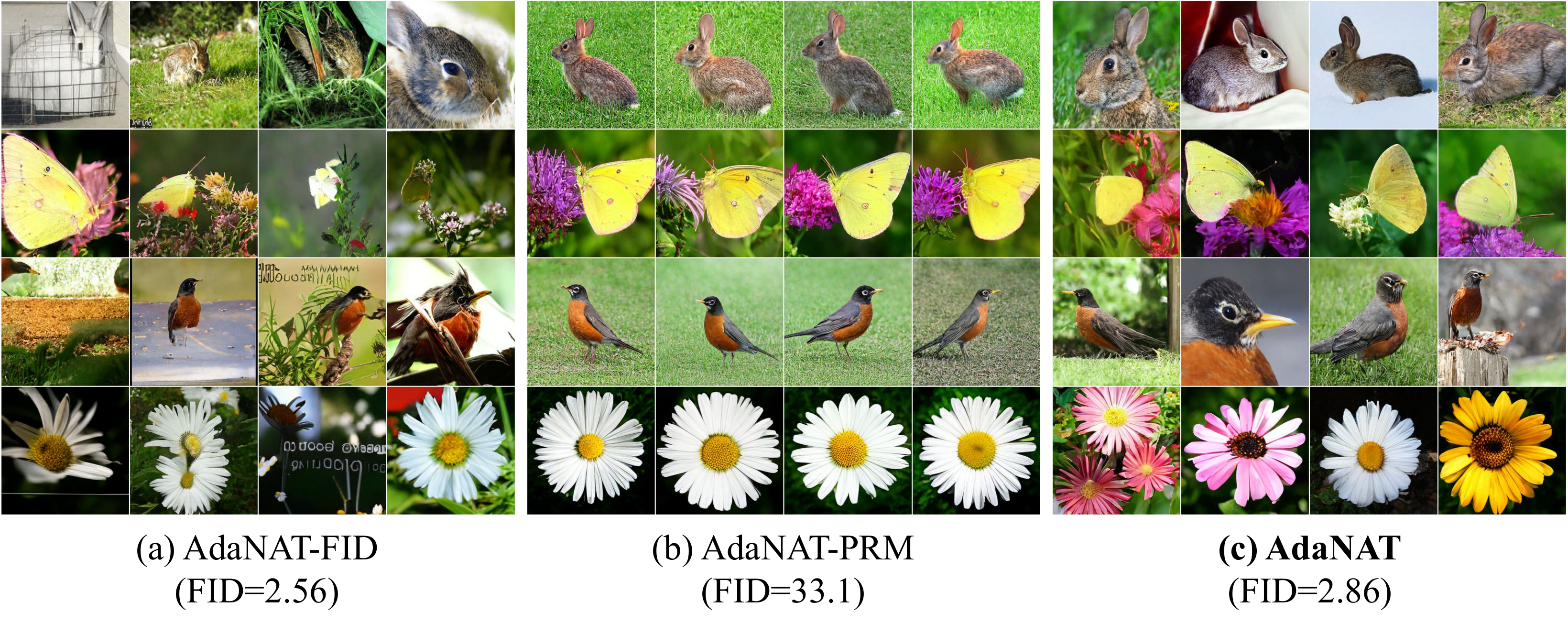}
\caption{
        \textbf{Ablation on different reward designs in \Ours.}
        (a) \Ours-FID: directly using FID~\cite{salimans2016improved} as the reward.
        (b) \Ours-PRM: using a pre-trained reward model~\cite{xu2024imagereward}.
        (c) \Ours: our main approach with adversarial reward model modeling.
}
\label{fig:comp_rewards}
\end{figure*}

\paragraph{Comparing different reward designs.}
In Figure~\ref{fig:comp_rewards}, we compare the generation results with different reward designs. Figure~\ref{fig:comp_rewards}a shows results\footnote{Initially, we tried learning the adaptive policy, but FID's inability to provide individual sample-wise reward signals led to convergence issues, prompting us to use a non-adaptive variant (see Appendix~\ref{sec:fid_reward}).} using Fréchet Inception Distance (FID)~\cite{salimans2016improved}. Despite this numerical superiority, we observe inconsistency between the low FID score and image quality, with many images displaying distortions or blurriness.
Figure~\ref{fig:comp_rewards}b depicts results from a pretrained reward model~\cite{xu2024imagereward}, which produced higher quality images but reduced diversity in poses, scales, and orientations. In contrast, Figure~\ref{fig:comp_rewards}c shows our adversarial reward model yielding high-quality images with a broader spectrum of variations, achieving a balanced trade-off between fidelity and diversity.

\section{Conclusion}

In this paper, we presented \Ours, a framework to enhance the generation policy of non-autoregressive Transformers. Our approach uses a policy network, trained via reinforcement learning within a Markov decision process, to adaptively select generation policies based on the sampling context. A key aspect of \Ours is the adversarial reward design, which overcomes limitations of straightforward reward signals, ensuring balanced quality and diversity in generated images. Validation on multiple benchmark datasets highlights \Ours's advantages in token-based image synthesis.

\section*{Acknowledgements}
This work is supported in part by the National Key R\&D Program of China under Grant 2021ZD0140407, the National Natural Science Foundation of China under Grants 62321005 and 42327901.

\bibliographystyle{splncs04}
\bibliography{main}

\clearpage
\section*{Appendix}
\appendix
\section{Implementation Details}
\label{sec:imp_details}
\subsection{Proximal Policy Optimization (PPO) Algorithm}
\label{sec:ppo}

Proximal Policy Optimization (PPO) algorithm offers a balance between sample efficiency and ease of implementation.
In this section, we elaborate in more detail of the PPO algorithm adopted in our paper.
First, consider a surrogate objective:
\begin{equation*}
    {L}^{\textnormal{CPI}}(\bm{\phi}) = \mathbb{E}_{t} \left[ \frac{\bm{\pi}_{\bm{\phi}}(\bm{a}_t|\bm{s}_t)}{\bm{\pi}_{\bm{\phi}_{\textnormal{old}}}(\bm{a}_t|\bm{s}_t)} \hat{A}_t \right],
    \label{eq:ppo_cpi}
\end{equation*}
where $\bm{\pi}_{\bm{\phi}}$ and $\bm{\pi}_{\bm{\phi}_{\textnormal{old}}}$ are the policy network before and after the update, respectively.
The advantage estimator $\hat{A}_t$ is computed by:
\begin{equation*}
    \hat{A}_t = -V(\bm{s}_t) + R(\bm{s}_T, \bm{a}_T),
\end{equation*}
where $V(\bm{s}_t)$ is a learned state-value function.
This objective function effectively maximizes the probability ratio $\rho_t({\bm{\phi}}) = \frac{\bm{\pi}_{\bm{\phi}}(\bm{a}_t \mid \bm{s}_t)}{\bm{\pi}_{{\bm{\phi}}_{\text{old}}}(\bm{a}_t \mid \bm{s}_t)}$ when considering the advantage of taking action $\bm{a}_t$ in state $\bm{s}_t$.
However, directly maximizing ${L}^{\textnormal{CPI}}(\bm{\phi})$ usually leads to an excessively large policy update, hence, we consider how to modify the objective, to penalize changes to the policy that move $\rho_t({\bm{\phi}})$ away from 1.
This gives rise to the clipped surrogate objective:
\begin{equation*}
    {L}^{\textnormal{CLIP}}(\bm{\phi}) = \mathbb{E}_{t} \Bigg[ \min\left( \rho_t({\bm{\phi}}) \hat{A}_t, \text{clip}\left(\rho_t({\bm{\phi}}), 1-\epsilon, 1+\epsilon\right) \hat{A}_t \right)\Bigg]
    \label{eq:ppo_clip}
\end{equation*}
where $\epsilon$ is a hyper-parameter that controls the range of the probability ratio.
Finally, to learn the state-value function, we additionally add a term for value function estimation error to the objective following~\cite{schulman2017proximal}, which results in the objective in Eq.~\textcolor{red}{11} in our main paper.

\subsection{Architecture}
Our policy network consists of a depth-wise convolution layer, a point-wise convolution layer and a multi-layer perceptron (MLP).
We take the NAT model's output feature as the generation status input, and additionally use adaptive layernorm (AdaLN)~\cite{perez2018film,peebles2023scalable} layers to incorporate timestep information into the policy network.
The architecture of the adversarial reward model follows the discriminator in StyleGAN-T~\cite{sauer2023stylegan}.
Notably, the policy network is highly lightweight, incurring \emph{negligible} additional inference cost:
{
\begin{center}
\tablestyle{0.8pt}{1}
\begin{tabular}{x{50}x{10}|x{90}x{80}x{110}}
    model & $T$ & infer. cost (all) & infer. cost (policy) & proportion (policy/all) \\\shline
    \Ours-S & 4 & 184.5 GFLOPs & 0.064 GFLOPs & 0.03\% \\
\end{tabular}
\end{center}
}

\subsection{Hyperparameter Details}
We perform the optimization loop of \Ours in Algorithm \textcolor{red}{1} for 1000 iterations.
In practice, we perform 5 gradient updates within each loop of Algorithm \textcolor{red}{1} for both the policy network and the adversarial reward model to facilitate a more stable and efficient optimization process in the minimax game.
For the optimization of the policy network, we adopt $\epsilon=0.2, c=0.5$ for the PPO objective, and use Adam~\cite{kingma2014adam} optimizer with a learning rate of $1\times10^{-5}$, $\beta_1=0.9, \beta_2=0.999$.
The batch size is set to 4096.
The $\sigma$ hyperparameter in Eq.~\textcolor{red}{9}, which balances exploration and exploitation, is set to 0.6 and reduced to 0.3 after 500 iterations.
For the adversarial reward model, we use Adam~\cite{kingma2014adam} optimizer with a learning rate of $1\times10^{-4}$ and $\beta_1=0.5, \beta_2=0.999$.
The batch size of updating the adversarial reward model is set to 1024 by default.
For experiments on ImageNet 512$\times$512~\cite{russakovsky2015imagenet} and CC3M~\cite{sharma2018conceptual}, we reduce the batch size to 512 to fit the memory constraints.
For the pre-training of our NAT models, we generally follow the training settings used in previous work~\cite{bao2022all}, with modifications on learning rate to 4e-4 and a larger batch size of 2048 on ImageNet dataset.
The results on CC3M is based on a publicly available  Muse model from github\footnote{https://github.com/baaivision/MUSE-Pytorch}.

\section{Experiment Details of FID-based Reward}
\label{sec:fid_reward}
When implementing FID-based reward design as described in Section \textcolor{red}{4.3}, we find that the FID-based reward model is not able to provide a stable and effective reward signal for the adaptive policy network, leading to divergence:
{
\begin{center}
        \tablestyle{2pt}{1}
    \begin{tabular}{x{90}x{50}x{50}|x{60}x{60}}
        && & \multicolumn{2}{c}{FID-50K$\downarrow$} \\
        dataset & model & $T$ & adaptive & non-adaptive \\\shline
        ImageNet 256 $\times$ 256 & \Ours-L & 8 & 55.4 (Fail) & \textbf{2.56}
    \end{tabular}
\end{center}
}
As a result, we adopted a non-adaptive version of policy network, where all samples share the same generation configuration.
Empirically, the non-adaptive policy network can also be optimized effectively with a low FID.
However, as discussed in Section \textcolor{red}{5.2}, this numerical superiority does not translate to a practical advantage in terms of sample quality.
The FID reward-based policy network fails to produce images of satisfactory quality.

\section{Practical Latency}
\label{sec:practical_latency}
\begin{figure*}[t!]
    \centering
    \includegraphics[width=.8\linewidth]{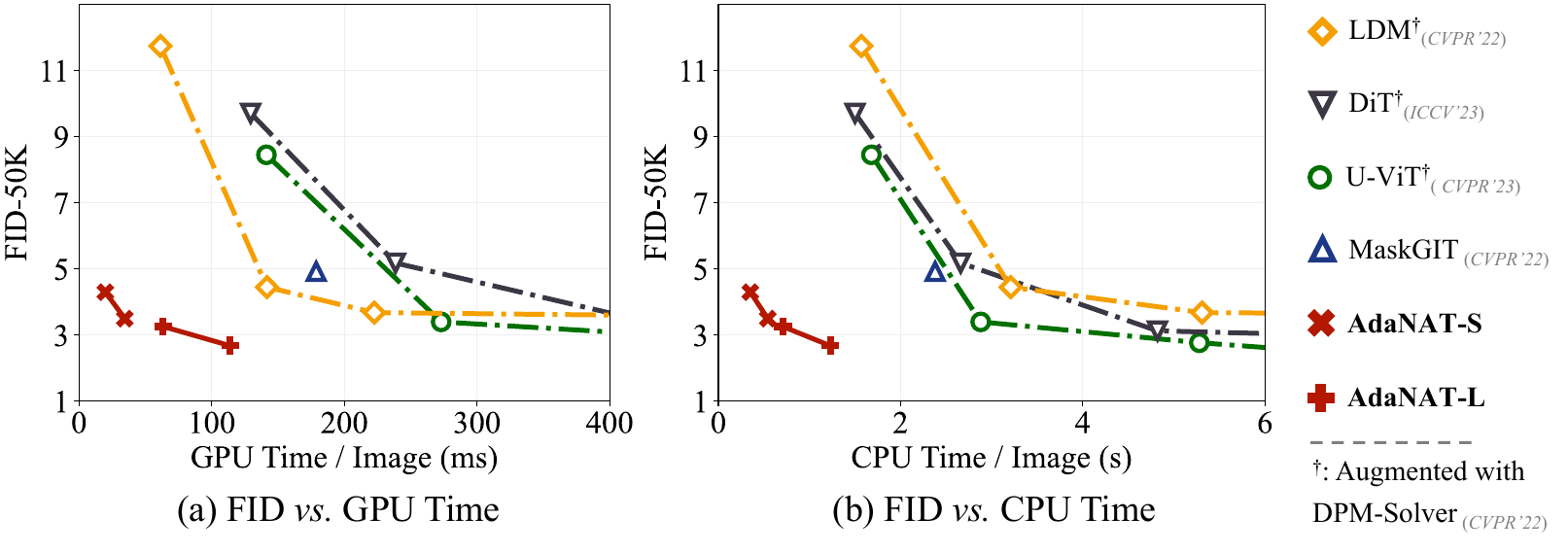}
        \caption{
        \textbf{Practical latency of \Ours on ImageNet 256$\times$256}.
        GPU time is measured on an A100 GPU with batch size 50. CPU time is measured on Xeon 8358 CPU with batch size 1.
        $\dagger$ : DPM-Solver~\cite{lu2022dpm} augmented diffusion models.
        \label{fig:practical_latency}
    }
\end{figure*}

Figure~\ref{fig:practical_latency} illustrates the comparison of the practical latency of \Ours against several competitive baselines on ImageNet 256$\times$256.
This comparison includes the latency on both GPU and CPU for generating a single image.
The results present a more comprehensive comparison on the efficiency \& efficacy tradeoff of \Ours and other methods in practical scenarios.

\begin{figure*}[t!]
    \centering
    \includegraphics[width=.7\linewidth]{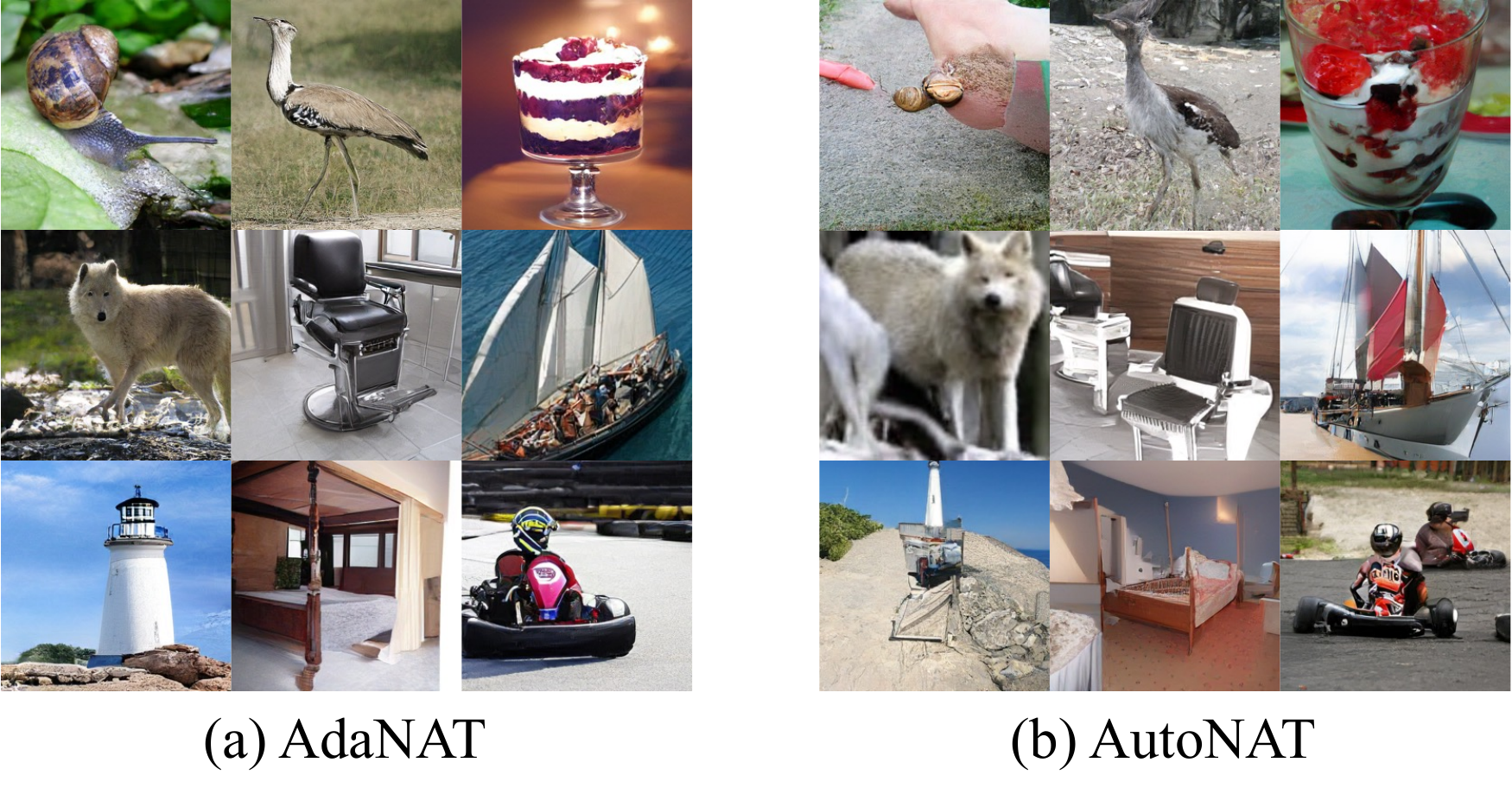}
    \caption{
        \textbf{Qualitative comparisons between \Ours and AutoNAT~\cite{ni2024revisiting} on ImageNet 256$\times$256}.
        \Ours generates images with superior visual quality.
        \label{fig:comp_autonat_ours}
    }
\end{figure*}
\section{Comparisons with AutoNAT}
\label{sec:adanat_vs_autonat}
Similar to \Ours, AutoNAT~\cite{ni2024revisiting} aims to enhance the policies in non-autoregressive Transformers.
It achieves this by optimizing a FID-based objective.
The table below provides quantitative comparisons between \Ours and AutoNAT on ImageNet 256$\times$256 with $T=8$.
    {
    \begin{center}
                \tablestyle{2pt}{1}
        \begin{tabular}{y{99}x{45}x{41}}
            method & TFLOPs$\downarrow$ & FID-50K$\downarrow$\\\shline
            AutoNAT-L~\cite{ni2024revisiting} & 0.9 & 2.68 \\
            AdaNAT-L (FID-based) & 0.9 & \textbf{2.56} \\
            AdaNAT-L (Adv-based) & 0.9 & 2.86
        \end{tabular}
    \end{center}
}
The results demonstrate that the FID-based approach achieves better quantitative metrics, with our FID-based AdaNAT model outperforming AutoNAT. However, as discussed in Section \textcolor{red}{5.2}, this optimization often leads to overfitting, resulting in subpar image quality. Consequently, we opted for an adversarial-based approach, which offers a more robust and favorable solution. Qualitative comparisons between AutoNAT and \Ours in Figure~\ref{fig:comp_autonat_ours} illustrate that \Ours generates images with superior visual quality.

\section{Potential Impact, Limitation, and Future Work}
\label{sec:impact}
As with any AI-generated content technology, there are potential ethical considerations and risks of misuse, such as creating misleading content, deepfakes, or spreading misinformation.
Additionally, like other data-driven approaches, the model may inadvertently reinforce biases present in the training data.
In terms of limitations and future work, it is essential to investigate the efficacy of \Ours on larger-scale datasets like laion-5B~\cite{schuhmann2022laion} and explore the performance of NAT models exceeding 1B parameters to understand scalability and robustness. Additionally, applying \Ours across more diverse generative tasks and domains~\cite{gal2022stylegan,guo2023smooth,guo2023zero,guo2024everything} could broaden its impact.
Integrating more advanced adaptive inference methods~\cite{yang2020resolution,yang2021condensenet,wang2021not,wang2020glance,huang2022glance,wang2021adaptive,wang2022adafocus,wang2022adafocusv3,zheng2023dynamic} and learning techniques~\cite{tang2020uncertainty,Wang2021RevisitingLS,ni2023deep,guo2024everything} can further enhance the capabilities and applicability of non-autoregressive Transformers.
Finally, better interpreting the decisions made by the policy network and translating them into insights for designing improved non-autoregressive transformer generation paradigms presents a valuable direction for future research.

\section{Scheduling Functions of Existing Works}
\label{sec:scheduling}
The scheduling functions of existing works~\cite{chang2022maskgit,li2023mage,chang2023muse,mentzer2023finite} are shown in the table below:

\begin{center}
\tablestyle{2pt}{1.14}
\begin{tabular}{y{88}|x{20}y{45}}
generation policy & \multicolumn{2}{c}{scheduling functions} \\\shline
re-masking ratio $m^{(t)}$     & \ \scriptsize  ${m{^{(t)}}}$ & \!\!\!\! \scriptsize $=\cos\frac{\pi (t+1)}{2T}$ \\
sampling temp. {$\tau_1^{(t)}$} & \ \scriptsize   ${ \tau_1{^{(t)}}} $ & \!\!\!\! \scriptsize $ = 1.0$ \\
re-masking temp.  $\tau_2^{(t)}$            & \ \scriptsize ${ \tau_2{^{(t)}}} $            & \!\!\!\! \scriptsize $= \frac{\lambda(T-t)}{T}$              \\
guidance scale  $w^{(t)}$            & \ \scriptsize  ${ w{^{(t)}}}$            & \!\!\!\! \scriptsize $=\frac{k(t+1)}{T}$              \\
\end{tabular}
\end{center}

\end{document}